\documentclass{article} 
\usepackage{iclr2021_conference,times}

\usepackage{amsmath,amsfonts,bm}

\def\eqref#1{equation~\ref{#1}}

\def\1{\bm{1}}

\def\eps{{\epsilon}}

\DeclareMathAlphabet{\mathsfit}{\encodingdefault}{\sfdefault}{m}{sl}
\SetMathAlphabet{\mathsfit}{bold}{\encodingdefault}{\sfdefault}{bx}{n}

\usepackage{hyperref}
\usepackage{url}
\usepackage[ruled,vlined]{algorithm2e}

\usepackage{graphicx}
\graphicspath{ {./image/} }
\title{FFPDG: Fast, Fair and Private Data Generation }

\author{Weijie Xu}

\begin{document}

\maketitle

\begin{abstract}
Generative modeling has been used frequently in synthetic data generation. Fairness and privacy are two big concerns for synthetic data. Although Recent GAN [\cite{goodfellow2014generative}] based methods show good results in preserving privacy, the generated data may be more biased. At the same time, these methods require high computation resources. In this work, we design a fast, fair, flexible and private data generation method. We show the effectiveness of our method theoretically and empirically. We show that models trained on data generated by the proposed method can perform well (in inference stage) on real application scenarios.


\end{abstract}

\section{Introduction}

Synthetic data [\cite{Rubin93}] is data that is artificially created rather than being generated by actual events. The availability of large synthetic data can bring many collaboration opportunities between related industries and research community. 
Synthetic data has been valuable to business functions, 
such as health [\cite{hwang2020eldersim}], robotics [\cite{duczek2021continual}], 
and financial services [\cite{Samuel19}]. 
However, personnel related synthetic data is rarely available online. [\cite{Rich2020}]

There are two concerns. 
First, any potential leakage of personal information can be harmful.[\cite{navaz2013human}] Recent work shows that machine learning models are highly susceptible to leak information from their training data [\cite{Reza17}]. If attackers have some real data, they are able to distinguish synthetic data from real data with high accuracy. [\cite{choquettechoo2021labelonly}] 
Second,  models trained on biased data have different predictive power across protected groups, such as race or gender. Bringing equal opportunity to protected groups is essential for synthetic data, as noted in [\cite{bolukbasi2016man} \cite{Joy18}].
Many methods [\cite{donini2020empirical} \cite{zafar2017fairness} \cite{JMLR:v20:18-262} \cite{hardt2016equality} \cite{woodworth2017learning} ]
have been provided to mitigate bias during model training or post model training. 
However,
[\cite{rachel19}] proved that there is no algorithm that is private, fair and better than a 
constant classifier. The proof relies on that test data distribution can be very different. Thus, we attempt to make datasets fair and private before training any machine learning methods. 

In this work, we propose a fast, fair, and private data generation method (FFPDG). Our algorithm can be used in supervised and unsupervised modeling applications. Our algorithm is fast since it has low run time complexity. We show that our algorithm has good fairness guarantee through experiments and mathematical proof. We show that our method preserves privacy under certain conditions. 


\section{Related Work}

\subsection{Private Data Generation}
[\cite{feldman2015certifying}] showed that removing personal information does not protect privacy. A privacy preserving dataset does not change outcome very much by inclusion or exclusion of a particular example. 

\textbf{Definition 1}(Differencial Privacy). A mechanism $A$ on a query functions f is $\epsilon$-differentially private if for all neighboring datasets $X$, $X^{'}$ which differ in a single record and for all possible measurement $S \subseteq R$, $\frac{Pr[A(f(X)) \in S]}{Pr[A(f(X')) \in S]} \leq exp(\epsilon)$ [\cite{cynthia06}]

Differentially Private Stochastic Gradient Descent (DP-SGD) [\cite{Abadi_2016}] is one of the first studies to make the Stochatic Gradient Descent computation differential private. It preserves differential privacy by clipping the gradient in the optimization's $l_2$ norm and adding noise. 

Private Aggregation of Teach Ensembles (PATE)[\cite{papernot2017semisupervised} \cite{papernot2018scalable}]deploy multiple teacher models which trained by disjoint datasets. PATE then deploy the teach models to unseen data to make predictions. For unseen data, the teacher models vote to determine the label combining random noise generated by laplace distribution. PATE further trains student model by only accessing to the privatized labels generated from the teacher's vote. Student cannot relearn an individual teacher's model as teacher is trained on disjoint datasets with laplacian noise. 
 
Generative Adversarial Network  and Variational Autoencoders\cite{kingma2014autoencoding} provide powerful methods to generate synthetic data using real data but they do not provide any privacy guarantee. These generation methods can combine with differentially private methods to achieve better privacy guarantee for generated data.

DPGAN [\cite{xie2018differentially}] proposes a framework for modifying the GAN [\cite{goodfellow2014generative}] framework to be differentially private, which relies on the PostProcessing Theorem [\cite{Cynthia14}] by learning a differentially private discriminator during training.   

PATE-GAN [\cite{yoon2018pategan}] modifies PATE framework to apply to GANs by applying PATE mechanism to the discriminator. The dataset is first partitioned into k subsets to train k teachers. Each teacher is trained by discrimination between generated data by generator and a disjoint subset of original data. The student discriminator is trained by voting from teachers result plus laplacian noise. At the end, the generator is trained to fool the student discriminator. 

Random OrthoNormal projection with GAUSSian generative model(RON-Gauss) [\cite{RChanyaswad19}] combines dimension reduction via random orthonormal projection and the Gaussian generative model for synthesizing differential private data. They are inspired by Diaconis-Freedman-Meckes (DFM) effect [\cite{Elizabeth12}] which shows that under suitable conditions, most projections of high dimensional data are nearly Gaussian. The data was generated by a process of normalization, random projection and gaussian modeling. Noise is added in the data normalization and gaussian model stages.

\subsection{Fair Processing}
We first give definition of Disparate Impact [\cite{Barocas18}], Equal Opportunity[\cite{heidari2018moral}] and Statistical Disparity. 

\textbf{Definition 2}(Disparate Impact). Given data set $D = (X, Y, C)$ with binary protected attribute C (e.g. race, sex, religion, etc, 0 is unprivileged group and 1 is privileged group), remaining attributes X and binary class to be predicted Y, we will say that C has disparate impact if $ \frac{Pr(Y = 1 | C = 0)}{Pr(Y = 1 | C = 1)} \leq 0.8$. 

\textbf{Definition 3}(Equal Opportunity/Equality of Odds) requires equal True Positive Rate(TPR) across subgroups: $P(Y^{'} = 1|Y = 1, C = 0) = P(Y^{'} = 1|Y = 1, C = 1)$ where $Y^{'}$ is the model output.

\textbf{Definition 4}(Statistical Parity) requires positive predictions to be unaffected by the value of the protected attribute, regardless of true label $P(Y^{'} = 1|C = 0) = P(Y^{'} = 1|C = 1)$

Another common definition is individual fairness, which means that people who are similar with respect to the task should be given similar predictions or decisions.  

\textbf{Definition 5}(Individual Fairness). IF $f : R^{n}  \to  R$ is a decision model, given appropriate distance functions - $d(.,.)$ on $R^{n}$ (the domain of f) and $D(.,.)$ on $R$ (the co-domain of f) - as well as threshold $\eps \geq 0$ and $\delta \geq 0$, the model is individually fair if, for any pair of inputs $x$, $x^{'}$ such that $d(x, x^{'}) \leq \eps$, we have $D(f(x), f(x')) \leq \delta$ [\cite{DBLP:journals/corr/abs-1104-3913}]

We can remove bias during data generation or during the downstream task. To address it from downstream tasks, there are two categories: 1. inprocessing methods [\cite{donini2020empirical} \cite{zafar2017fairness} \cite{JMLR:v20:18-262} \cite{perrone2021fair}] that change the objective function optimized during training to include fairness constraints and 2. post-processing [\cite{hardt2016equality} \cite{woodworth2017learning}] methods that modify the outcome of the existing models by changing decision boundary. Some of these method such as [\cite{DBLP:journals/corr/abs-1812-02696}] preserve privacy and fairness. Our work focuses on removing bias during data generation process and stronger privacy assumption.  

Disparate impact remover (DIR) [\cite{feldman2015certifying}] creates a distribution for each protected attribute. DIR then creates a median distribution over all distribution. The outcome variable is then projected by median distribution. This method is rank preserving for samples from each protected group. 

Fair Max Entropy Distributions (FairMaxEnt) [\cite{pmlr-v119-celis20a}] is a maximum entropy based approach to eliminate bias from data. The method uses priors and marginal vector to constrain on statistical rate [\cite{feldman2015certifying}] and representation rate [\cite{hardt2016equality}] of the generated data. Then, it solves dual of the max-entrophy framework to optimize data distribution. This method is faster than methods such as [\cite{calmon2017optimized}]. However, this method does not consider individual fairness and only works for binary data.

\section{Methodology}

In our proposed method, we first use FairMaxEnt to create unbiased data and then use RON-Gauss to generate private data.(See Algorithm \ref{algo1})

For step 2, we use discretization methods suggested in the original paper from [\cite{pmlr-v119-celis20a}]. For step 1 and 3, we use a dictionary to store the mappings from binary data to original data. For step 4,we use the same preprocessing methods described in [\cite{xu2018synthesizing}] for categorical features. After data generation, we map binary features to original datasets by uniformly sampling from mappings. Since FairMaxEnt is a distribution over the domain rather than reweighting, some generated data does not have mapping to original datasets. We then sample data from the closest mappings. For step 7, we randomly sample matrix and use QR factorization [\cite{trefethen97}] to get W. For step 8, we use normal distribution as generative model for unsupervised and continuous outcome variable. We can use Gaussian Mixture Model [\cite{Amendola_2016}] to generate data for categorical outcome variable. Thus, our method is flexible since it can be used in regression, classification and unsupervised settings. For step 9, our study uses training data percentile as threshold to map continuous data back to binary/categorical features and  we also restrict continuous data to range of the minimum and maximum from the original data.

\begin{algorithm}[H]
 \KwData{dataset $X \in R^{d\times n}$, dimensions $p < d$, and $\epsilon_{\mu}, \epsilon_{\sum} > 0$}
 \KwResult{$x_{1}^{DP}$, ....  $x_{n'}^{DP}$}
 Step 1: Create a dictionary D to map data X to binary data B\;
 Step 2: Obtain the fair processed data $B^{'}$ from FairMaxEnt with inputs B\;
 Step 3: Map $B^{'}$ back to X using nearest neighbor in D and uniform sampling\;
 Step 4: Pre-normalize: change categorical feature to one hot encoding plus white noise and $x_{i} := \frac{x_{i}}{\| x_{i} \|}$ for all $x_i \in X$\;
 Step 5: Center the data: $\bar{X} = X - \mu^{DP} 1^{T} $ where $\mu^{DP} = (\frac{1}{n} \sum_{i=1}^{n} x_i) + Z$ and $z_{j}(i)$ is drawn i.i.d from $Lap(2 \sqrt{d} / n \epsilon{\mu})$ \;
 Step 6: Re-normalize: $\bar{x_i} = \bar{x_i} / \| \bar{x_i} \|$ for all $\bar{x_i} \in \bar{X}$.\;
 Step 7: Construct a RON projection matrix $W = [q_{1}, ... q_{p}] \in R^{d \times p}$ and project the data $X' = W^{T} \bar{X} \in R^{p \times n}$.\;
 Step 8: Draw synthetic data $x_{i}^{DP} \in R^{p}$ from $N(0, \sum^{DP}) $ where $\sum^{DP} = (\frac{1}{n} X' X'^{T}) + Z$ and $z_{j}(i)$ is drawn i.i.d from $lap(2\sqrt{p}/n \epsilon_{\sum})$.\;
 Step 9: Project it back to original space using W and transform it to original data format.

 \caption{Fast, Fair and Private Data Generation}
 \label{algo1}
\end{algorithm}

The proposed method is fast. Suppose d is the dimension of data and n is the number of samples in the data and the number of data FFPDG generates. Step 1 runs in time $nd$. Step 2 runs in time polynomial in d, n and the bit complexity of the  input parameters. (See Appendix D.3 in [\cite{pmlr-v119-celis20a}]). Step 3 runs in time $n^{2}$ as identifying nearest neighbor runs in time N at most.  Step 4 to 6 runs in time $n d$. Steps 7 runs in time $ n d^{2}$ as projected dimension is smaller than real dimension and QR decomposition runs in $d^{3}$. Steps 8 runs in time $n^{2} d$ and step 9 runs in time $n$. Thus, the whole algorithm runs in time polynomial in $d^{3}$, $n^{2}$ and the bit complexity of the  input parameters. If we want to generate data from extreme large data set, this method is faster and more computationally efficient than GAN based methods.

The proposed method is differentially private under some conditions. For step 1-4, data is generated by using KL divergence [\cite{shlens2014notes}] framework. The resulting posterior is robust in terms of KL-divergence to small changes in the data.(See Theorem 1 in Appendix). The generated samples are differential private under lipschitz continuity assumption.(See Theorem 2 in Appendix). Thus, till this step, our method is 2L-differentially private where L is mentioned in Appendix B Assumption 1. For step 5-8, it is approved by [\cite{RChanyaswad19}] that it is $(\epsilon_{\mu} + \epsilon_{\sum})$-differentially private. According to serial composition theorem from [\cite{cynthia08}], if we can find L, our method is $(2L + \epsilon_{\mu} + \epsilon_{\sum})$-differentially private. We inject noise in different part of process such as sampling, data normalizing and data generation. Even if attackers know the whole generation process, it is almost impossible for them to estimate most parameters accurately. 

Our method preserves group fairness. FairMaxEnt preserve both fairness constraints. While a few studies [\cite{cynthia13} \cite{michael18} \cite{satyal19}] argue that differentially private algorithms make unfair decisions, the datasets used in their study are not fair. We do not find evidence that shows that differentially private algorithm make unfair decision on fair dataset. 

Our method preserves individual fairness with enough samples in regression settings .  For step 1-4, we only sample data from previous distribution which preserves individual fairness. For step 5-7, after normalization and RON projection, we prove that the distance between different data point is getting closer.  (See Theorem 3 in Appendix for proof) For step 8, we show that when outcome variable is generated by linear combination of feature space plus noise, sample generated from this model converge to data generation space as sample size goes to infinity. (See Theorem 4 in Appendix for proof). To summarize, our generated method only generate data that is closed to original data space.

\section{Metric and Experiment}

\subsection{Metric}
AUCROC[\cite{yoon2018pategan}]: We train few models on generated data. Then, we test those models on real data and use area under the receiver operating characteristics curve(AUCROC) as our metric. If we saw high AUCROC on the real data for models that were trained on synthetic data, we can infer that synthetic data has captured the relationship between features and labels well. These synthetic data can be used to train models without ever seeing the real data. To have a stable result, we choose best AUCROC performance among models that include logistic regression, Random Forest[\cite{breiman01}], Neural Network, Gaussian Naive Bayes[\cite{Rish01anempirical}],  Gradient Boosting Classifier[\cite{jer01}], Bernoulli Naive Bayes, Decision Tree and Linear Discriminant Analysis.

DEO/DSP: $Y^{'}$ is the predicted value of outcome variable. Difference in equal opportunity DEO  ($|P(Y^{'} = 1|Y = 1, C = 0) - P(Y^{'} = 1|Y = 1, C = 1)|$)  and Difference in statistical parity DSP ($|P(Y^{'} = 1|C = 0) - P(Y^{'} = 1|C = 1)|$) (Note that DSP also called Demographics Parity Difference (DPD) [\cite{zafar2017fairness}]) [\cite{perrone2021fair}] are two measures we use to evaluate group fairness of the generated result. We use logistic regression ,  Random Forest, Neural Network, Gaussian Naive Bayes,  Gradient Boosting Classifier, Bernoulli Naive Bayes, Decision Tree\cite{ross86} and Linear Discriminant Analysis and choose the average of them as default models to predict $Y^{bar}$.  (For other metric, you can look at [\cite{DBLP:journals/corr/abs-1104-3913}] )

LRD: We build a logistic regression classifier that learns to tell the synthetic data apart from the real data (LRD), which later on is evaluated using cross validation.  The output of the metric is one minus the average AUCROC score obtained. The higher the score, the harder it is for the model to distinguish real data from synthetic data. We use this metric to understand if the generated data is robust against adversarial attacks such as membership inference attack.

\subsection{Hyperparameter Tuning}

 We set epsilon equals to $1$ for differential privacy. We choose repair level for DIR equal to 1. We choose smooth factor equals to 0.1 and error parameter equal to 0 for FairMaxEnt. We choose $\sigma$ equals to 2, clip coefficient equals to 0.1, micro batch size equals to 8, number of epochs equals to 500 and batch size equals to 64 for DPGAN. Number of teacher equals to 10, teacher and student iterations equal 5, number of moments equal to 100 for PATEGAN. For RON-Gauss, we set $\epsilon_{\mu}$ to be 30 percent of $\epsilon$ and $\epsilon_{\sum}$ to be 70 percent of $\epsilon$. All experiments are run on AWS with ml.p2.16xlarge.

\subsection{Experiment}

The Adult [\cite{Dua:2019}] dataset contains demographic information of individuals along with a binary label of whether their annual income is greater than  50k. In our analysis we include attributes race (white VS non white), sex, age and education years. We use gender as the protected attribute.

The COMPAS [\cite{barenstein2019propublicas}] dataset contains information on criminal defendants at the time of trial, along with post-trail instances of recidivism. We include attributes such as sex, race, priors count and charge degree as features. We use gender as the protected attribute.

We first study whether we should put fair processing first or private data generation first. Our finding is that if we put fair processing after private data generation, it becomes harder to decrease DSP and increase AUCROC. Although more experiments may be needed to see if this result can be generalized, we recommend that fair processing occurs before private data generation.  Private data generation relies on adding noise in labeling or training process. If data has many outliers, this private generated data may exaggerate these outliers and generate more noisy data. Fair processing method such as FairMaxEnt can mitigate the influence of outliers by projecting feature space to binary and fitting a distribution. 

\begin{table*}[ht]
\centering

\begin{tabular}{c|c|c|c|}
\hline
\multicolumn{1}{c}{AUCROC} \vline
& \multicolumn{1}{c}{DP WGAN} \vline
& \multicolumn{1}{c}{PATE GAN} \vline
& \multicolumn{1}{c}{RON GAUSS} \\
\hline

FairMaxEnt First  & 0.62 & 0.75 & 0.76 \\ 

\hline
FairMaxEnt Last  & 0.53 & 0.72 & 0.74 \\

\hline
\end{tabular}

\caption{Compare pre/post fair processing influence AUCROC}
\label{table3}
\end{table*}

\begin{table*}[ht]
\centering

\begin{tabular}{c|c|c|c|}
\hline
\multicolumn{1}{c}{DSP} \vline
& \multicolumn{1}{c}{DP WGAN} \vline
& \multicolumn{1}{c}{PATE GAN} \vline
& \multicolumn{1}{c}{RON GAUSS} \\
\hline

FairMaxEnt First  & 0.07 & 0.11 & 0.07 \\ 

\hline
FairMaxEnt Last  & 0.22 & 0.16 & 0.40 \\

\hline
\end{tabular}

\caption{Compare pre/post fair processing influence DSP}

\end{table*}

Then, we compare our result with combinations of reweight[\cite{kamiran2012}]/DIR/FairMaxEnt as fair processing and DPWGAN/PATE-GAN/RON-Gauss as private data generation methods. Code we used: [ \cite{bellamy2018ai} \cite{vijay20} \cite{7796926}]

We also compare the speed of our method with others. We choose the fastest DPWGAN and PATEGAN based method. We then compare them with FFPDG on COMPAS and Adult dataset. We can see that FFPDG is much faster than GAN based methods.

\begin{table*}[ht]
\centering

\begin{tabular}{c|c|c|c|}
\hline
\multicolumn{1}{c}{Dataset} \vline
& \multicolumn{1}{c}{FairMaxEnt + DPWGAN} \vline
& \multicolumn{1}{c}{reweight + PATEGAN} \vline
& \multicolumn{1}{c}{FFPDG} \\
\hline

COMPAS  & 36 seconds & 29 minutes & 1 seconds \\ 

\hline
Adult  & 47 seconds & 29 minutes & 1 seconds \\

\hline
\end{tabular}

\caption{Compare pre/post fair processing influence DSP}

\end{table*}

Based on our experiment, our proposed method achieves good AUCROC while maintains the highest LRD and lowest DSP. (See table 3 and 4 in appendix for full results) We run the same experiment on COMPAS dataset[\cite{barenstein2019propublicas}] and see similar result but with relatively lower AUCROC. High AUCROC means data generated by our method has good predictability power on real data. High LRD means that our generated data is hard to distinguish from real data. Low DSP shows that our method preserves group fairness. In general, FairMaxEnt performs better in DSP compare to all other fair processing methods. RON-Gauss performs better in LRD compare to DPWGAN and PATEGAN. 





\begin{figure}[h]

\includegraphics[width=7cm, height=4cm]{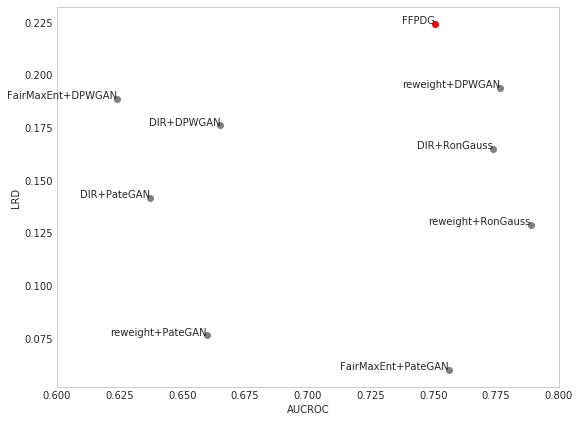}
\includegraphics[width=7cm, height=4cm]{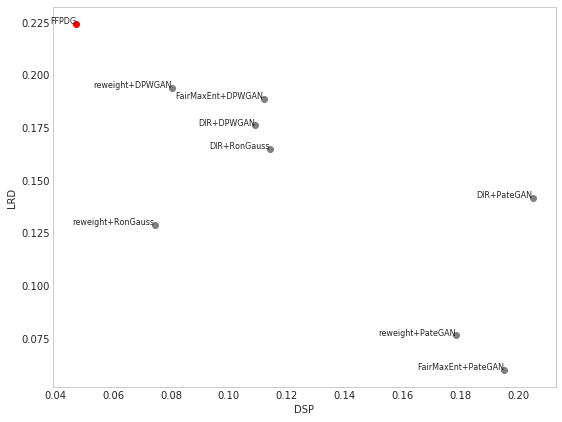}
\caption{Comparison of different methods in Adult data set}
\end{figure}

\begin{figure}[h]

\includegraphics[width=7cm, height=4cm]{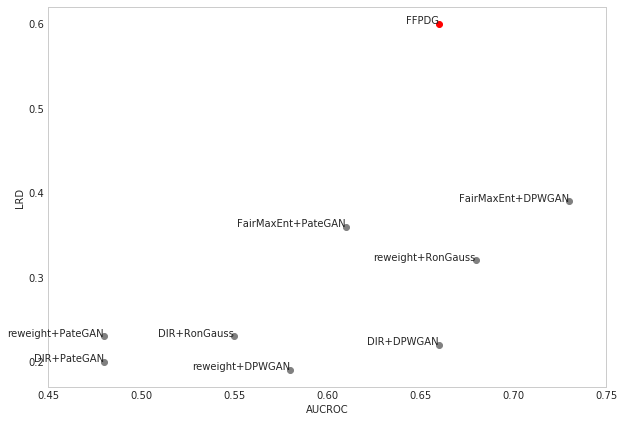}
\includegraphics[width=7cm, height=4cm]{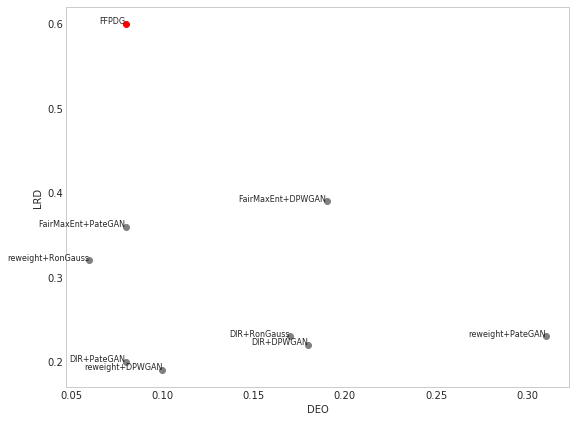}
\caption{Comparison of different methods in COMPAS data set}
\end{figure}

\section{Limitations and Conclusions}


\subsection{Limitations}
Although, DFM shows that data can be projected to low dimensional Gaussian distribution. It becomes harder to find good projection with categorical data and increase of dimensions. We see the predictability on real data decreases as dimensions or categorical features increase. We may need to change step 8 in our proposed method for better data generation on high dimensional data. 

LRD is very low for almost all private preserved generated data. We further study methods such as CTGAN[\cite{DBLP:journals/corr/abs-1907-00503}] and Copula GAN[\cite{kamthe2021copula}] using Adult dataset to see if this result can be generalized for other data generation technique. The result shows that for the same procedure, the LRD for them are 0.6 and 0.79. However, if we use the same post processing such as capping by max and min and changing continous data back to binary/categorical features. The LRD drops to 0.19 and 0.15. To make our method scale, better postprocessing method for categorical feature is needed.

Our method does not inference well on dataset that is biased. Our method removes biases from the dataset. If test data is hugely biased, distribution shift may occur between generated data and test data. Model trained on these generated data may not inference well on biased test data.

\subsection{Conclusion}

To conclude, we design a method to generate fair and unbiased data. The method of data generation has the following benefits: It is flexible and fast, the data created can produce models that perform well in real datasets, robust to membership inference attack and preserve individual/group fairness. 

There are few directions of future study: (1) Make our algorithm scale to high dimensional datasets. (2) Find a better method to process categorical features. (3) Merge fair processing and private data generation into one step.

\bibliography{iclr2021_conference}
\bibliographystyle{iclr2021_conference}

\appendix
\section{Tables}

\subsection{Results from COMPAS}

\begin{table*}[ht]
\centering

\begin{tabular}{c|c|c|c|c|c}
\hline
\multicolumn{1}{c}{} \vline
& \multicolumn{1}{c}{Metric} \vline
& \multicolumn{1}{c}{Original Data} \vline
& \multicolumn{1}{c}{DPWGAN} \vline
& \multicolumn{1}{c}{PATEGAN} \vline
& \multicolumn{1}{c}{RON GAUSS} \\
\hline

No Preprocessing & AUCROC  & 0.84 & 0.52 & 0.75 & 0.71 \\ 
No Preprocessing & DEO  &0.39 & 0.24 & 0.16 & 0.51 \\
No Preprocessing & DSP  &0.13 & 0.12 & 0.08 & 0.21 \\
No Preprocessing & LRD  &0.99 & 0.16 & 0.12 & 0.18 \\
\hline
REWEIGHT & AUCROC  & 0.83 & \textbf{0.78} & 0.65 & \textbf{0.79} \\ 
REWEIGHT & DEO  & 0.41 & 0.20 & \textbf{0.01} & 0.16 \\
REWEIGHT & DSP  & 0.13 & 0.08 & 0.18 & 0.07 \\
REWEIGHT & LRD  & 0.99 & 0.19 & 0.08 & 0.13 \\
\hline
DIR & AUCROC  & 0.83 & 0.67 & 0.64 & \textbf{0.77} \\
DIR & DEO  & 0.41 & 0.02 & \textbf{0.01} & 0.32 \\
DIR & DSP  & 0.13 & 0.11 & 0.20 & 0.11\\
DIR & LRD  & 0.99 & 0.18 & 0.14 & 0.16 \\
\hline
FairMaxEnt & AUCROC  & 0.81 & 0.62 & 0.75 & \textbf{0.76} \\
FairMaxEnt & DEO  & 0.06 & 0.07 & 0.12 & 0.07 \\
FairMaxEnt & DSP  & 0.12 & 0.07 & 0.20 & \textbf{0.05} \\
FairMaxEnt & LRD  & 0.82 & 0.19 & 0.06 & \textbf{0.22} \\

\hline
\end{tabular}

\caption{Model performance with different metric for combination of fair preprocessing and private data generation for Adult data}

\end{table*}

\begin{table*}[ht]
\centering

\begin{tabular}{c|c|c|c|c|c}
\hline
\multicolumn{1}{c}{} \vline
& \multicolumn{1}{c}{Metric} \vline
& \multicolumn{1}{c}{Original Data} \vline
& \multicolumn{1}{c}{DPWGAN} \vline
& \multicolumn{1}{c}{PATEGAN} \vline
& \multicolumn{1}{c}{RON GAUSS} \\
\hline

No Preprocessing & AUCROC  & 0.74 & 0.63 & 0.73 & 0.66 \\ 
No Preprocessing & DEO  &0.38 & 0.09 & 0.06 & 0.08 \\
No Preprocessing & DSP  &0.01 & 0.04 & 0.04 & 0.05 \\
No Preprocessing & LRD  &0.99 & 0.33 & 0.33 & 0.24 \\
\hline
REWEIGHT & AUCROC  & 0.73 & 0.58 & 0.48 & 0.68 \\ 
REWEIGHT & DEO  & 0.38 & 0.10 & 0.31 & 0.06 \\
REWEIGHT & DSP  & 0.01 & 0.07 & 0.05 & 0.02 \\
REWEIGHT & LRD  & 0.98 & 0.19 & 0.23 & 0.32 \\
\hline
DIR & AUCROC  & 0.73 & 0.66 & 0.48 & 0.55 \\
DIR & DEO  & 0.36 & 0.18 & 0.08 & 0.17 \\
DIR & DSP  & 0.01 & 0.06 & 0.04 & 0.04 \\
DIR & LRD  & 0.89 & 0.22 & 0.20 & 0.23 \\
\hline
FairMaxEnt & AUCROC  & 0.73 & 0.73 & 0.61 & 0.66 \\
FairMaxEnt & DEO  & 0.06 & 0.19 & 0.08 & 0.08 \\
FairMaxEnt & DSP  & 0.02 & 0.02 & 0.05 & 0.04 \\
FairMaxEnt & LRD  & 0.68 & 0.39 & 0.36 & \textbf{0.60} \\

\hline
\end{tabular}

\caption{Model performance with different metric for combination of fair preprocessing and private data generation for COMPAS data}
\label{table2}
\end{table*}

\section{Theorem}

\textbf{Assumption 1} (Individual Bias) $f : R^{n}  \to  R$ said to be individually biased if there exists a pair of valid inputs $x$ and $x^{'}$, with $|f(x) - f(x^{'})| > \delta$, such that
$|x_{i} - x_{i}^{'} | \leq \epsilon_{j} $for all $i \in S_{j}$ , and for all $ j = 1, . . . , t$.
Such a pair $(x, x^{'})$ is called an individual bias instance of the model $f$. [\cite{john2020verifying} \cite{kleinberg2016inherent}]

\textbf{Assumption 1} (Lipschitz continuity) Let $f(x, \theta) = \ln{ p_{\theta} (x)}$ be the log probability of x under $\theta$. $\rho : S \times S \rightarrow R_{+}$ is the pesudo distance metric. The Lipschitz constant for a parameter value $\theta$ is $ l(\theta) = inf \{\mu : |f(x, \theta) - f(y, \theta) | \leq \mu \rho(x,y) \forall x, y \in S  \}$. We assume there exist some $L < \infty$ such that: $$
l(\theta) \leq L, \theta \in \Theta $$ [\cite{JMLR:v18:15-257}]

\textbf{Theorem 1} When $\xi$ is a prior distribution on $\Theta $ and $\xi(\cdot| x)$ and $\xi(\cdot| y)$ are the respective posterior distribution for data sets $x, y \in S $, under a peseudi-metric $\rho$ and $L > 0$ satisfying Assumption 1, 
$$D(\xi(\cdot| x) \|  \xi(\cdot| y)) \leq 2L \rho(x, y)$$ [\cite{JMLR:v18:15-257}]

\textbf{Theorem 2} Under a pseudo-metric $\rho$ and $L > 0$ satisfying Assumption 1, for all $x, y \in S$, $B \in \sigma_{\theta} $:
$$\xi(B| x) \leq exp\{2L \rho(x,y) \} \xi(B|y)$$ (This means posterior $\xi$ is  $(2L, 0)$-differentially private under pseudo-metric $\rho$) [\cite{JMLR:v18:15-257}]

\textbf{Theorem 3} for $\forall x_{1}, x_{2} \in D$, if $z_{i} = z_{j}$,  then $\| f(x_{i}) - f(x_{j}) \|_{F} < \| x_{i} - x_{j} \|_{F}  $ where $z_i, z_j$ is noise drawn from laplacian noise and f represent step 5-7. W is ron projection and $\mu$ is the average from normalization.

Proof: Since we assume $z_{i} = z_{j}$, $u_{i} = \frac{1}{n}\sum_{i=1}^{n} (x_i) + z_i = \frac{1}{n}\sum_{i=1}^{n} (x_i) + z_j = u_{j}$
$$\| f(x_{i}) - f(x_{j}) \|_{F}  = \| W^{T}(x_{i} - \mu_i) -  W^{T}(x_{j} - \mu_j) \|_{F}$$

$$=\| W^{T}(x_{i} - x_{j} )\|_{F} $$
$$= \sqrt{tr((x_{i} - x_{j} )^{T} WW^{T} (x_{i} - x_{j} ))}$$
    $$ = \sqrt{tr((x_{i} - x_{j} )^{T} WW^{T} WW^{T} (x_{i} - x_{j} ))}  $$                [This is because $W^{T} W = I$]
    $$ = \| W W^{T}(x_{i} - x_{j} )\|_{F} $$ 
    [Let $P = W W^{T}$]
    $$ = \| P(x_{i} - x_{j} )\|_{F} $$ 
    $$ = <P(x_{i} - x_{j} ), P(x_{i} - x_{j} )>_{F}$$
    $$ = <P(x_{i} - x_{j} ), (x_{i} - x_{j} )>_{F}$$
    [Consider $x' = x - Px$, $<Px, x>_{F} = <Px, x' + Px> = <Px, Px>_{F} + <Px, x'>_{F} = <Px, Px>_{F}$ since $Px \perp x'$, $<Px, x'>_{F} =  0$ ]
    
    On the other hand, we have:
    $ \| P(x_{i} - x_{j} )\|_{F}^{2} = <P(x_{i} - x_{j} ), (x_{i} - x_{j} )>_{F}^{2} \leq \| P(x_{i} - x_{j} )\|_{F}  \| (x_{i} - x_{j} )\|_{F}$ because of Cauchy-Schwarz inequality. This means $ \| P(x_{i} - x_{j} )\|_{F} <  \| (x_{i} - x_{j} )\|_{F}$
    Thus we have $\| f(x_{i}) - f(x_{j}) \|_{F} = \| P(x_{i} - x_{j} )\|_{F}  <  \| (x_{i} - x_{j} )\|_{F}$

\textbf{Theorem 4} Suppose $Y = X \beta + \epsilon$, in regression settings, 
\[
\begin{bmatrix}
x_{i}\\
y_{i}
\end{bmatrix} \in  R^{p+1}
\] 
by drawing samples from 
\[
N\left(0, \begin{bmatrix}
\frac{1}{n} (XX^{T} + Z_{x}) & \frac{1}{n}X^{T}Y\\
\frac{1}{n}Y^{T} X &  \frac{1}{n}(Y Y^{T} + z_{y})
\end{bmatrix}\right).
\]  
Here, $\| x_{i} \| = 1$.
For simplicity, we assume $Z_{x}$ is positive semidefinite diagnol matrix. Any none zero element in Z iid drawn from $Lap((2\sqrt{p} + 4a\sqrt{p} + a^{2})/\sqrt{n}\epsilon_{\sum})$. Then, the variance of y is less than $\frac{1}{n} \epsilon^{T} \epsilon +  \frac{1}{n} \beta^{-1} Z_{X} \beta + z_{Y}$ and the distance between y and $X \beta $ is less than $z_{max} \beta^{T} (X X^{T})^{-1} $ where $z_{max}$ is the largest absolute value in diagonal matrix Z

Proof:
$$Avg(y|x) = \frac{1}{n} Y^{T} X(\frac{1}{n} X X^{T} + Z_{x})^{-1} x$$
[$(A + B)^{-1} = A^{-1} - A^{-1} B (A + B)^{-1}$]
$$= Y^{T} X((X X^{T})^{-1} - (X X^{T})^{-1}  Z_{x}(X X^{T} + Z_{x})^{-1}) x $$
$$\leq  Y^{T} X((X X^{T})^{-1} - (X X^{T})^{-1}  Z_{x}(X X^{T})^{-1}) x$$
$$=  Y^{T} X(X X^{T})^{-1}x    - Y^{T} X(X X^{T})^{-1}  Z_{x}(X X^{T})^{-1} x$$
[$\beta = (X^{T} X)^{-1} X^{T} Y$ and $((X^{T} X)^{-1})^{T} =  (X^{T} X)^{-1}$]
$$= \beta^{T} x - \beta^{T}  Z_{x}(X X^{T})^{-1} x $$
Since $\| x_{i} \| = 1$ and $Z_{x}$ is a diagonal matrix , $Avg(y|x)  - \beta^{T} x  \leq z_{max} \beta^{T} (X X^{T})^{-1} $

$$Var(y|x) = \frac{1}{n} (Y Y^{T}  - Y^{T} X(X^{T} X + Z_{X})^{-1} ( X^{T} Y) )+ z_{Y} $$
If $(x, y)$ is a multivariate Gaussian, where the mean is 0 and the variance is 
\[
\begin{bmatrix}
\sum_{11} & \sum_{12}\\
\sum_{21} & \sum_{22} 
\end{bmatrix},
\] 
then $Var(y|x) = \sum_{22} - \sum_{21} \sum_{11}^{-1} \sum_{12}$.
$$( X^{T} X + Z_{x})^{-1} = ( X^{T} X)^{-1} - ( X^{T} X)^{-1} Z_{x} (X^{T} X + Z_{x})^{-1}$$
$$\geq ( X^{T} X)^{-1} - ( X^{T} X)^{-1} Z_{x} (X^{T} X )^{-1}$$

$$\frac{1}{n} (Y^{T} Y  - Y^{T} X ( X X^{T} + Z_{X})^{-1} X^{T} Y )+ z_{Y} $$
$$ \leq \frac{1}{n} (Y^{T} Y  - Y^{T} X  (( X X^{T})^{-1} - ( X X^{T})^{-1} Z_{x} ( X X^{T} )^{-1})  X^{T} Y + z_{y}\textbf{}$$
[$H = X (X^{T} X)^{-1} X^{T}$ and H is symmetric and $\beta = (X^{T} X)^{-1} X^{T} Y$]
$$= \frac{1}{n}(Y^{T} Y - Y^{T} H Y  +  \beta^{-1} Z_{X} \beta) + z_{Y} $$
[Since $H*H = H$, we have $I - H = I - HH = I - H - H + HH = (I - H) (I - H)$ since I - H is symmetric, $I - H = (I - H)^{T} (I - H)$]
$$= \frac{1}{n} (Y^{T} (I - H )^{T} (I - H) Y +  \beta^{-1} Z_{X} \beta) + z_{Y}$$
$$= \frac{1}{n} \epsilon^{T} \epsilon +  \frac{1}{n} \beta^{-1} Z_{X} \beta + z_{Y}$$

Thus, as $n -> \infty$, $z_{max}$ and  $z_{y}$ go to 0, both difference and variance goes to 0. $y|x$ convergence to $\beta^{T} x$ in probability. 

For the same setting as Theorem 4, our proposed method is not individually biased under definition 1 as sample goes to infinity. y|x is sampled from $N(\beta^{T}x + \epsilon, \sigma)$, where $\epsilon$ and $\sigma$ go to 0. Thus, for any samples such that $|x_{i} - x_{i}^{'} | \leq \epsilon_{j} $, we have $|f(x) - f(x^{'}) | \to \beta^{T} |x-x'|\leq \beta_{max} \sum_{j = 1}^{t} \epsilon_{j}$  where $\beta_{max}$ is the largest absolute value in $\beta$. We can find constant C, as sample size goes to infinitiy, $|f(x) - f(x^{'}) | \leq C +  \beta_{max} \sum_{j = 1}^{t} \epsilon_{j}$. Thus, the difference of outcome variable is bounded. Our analysis is based on definition 1. It may not hold in other definitions of individual bias.[\cite{dwork2011fairness} \cite{DBLP:journals/corr/abs-1806-06122} \cite{kearns2019average}]


\end{document}